\documentclass[sigconf, screen, nonacm]{acmart}
\usepackage{multirow}
\usepackage{colortbl}
\usepackage{xcolor}
\usepackage{balance}

\AtBeginDocument{%
  }

\newcommand{\grayunderline}[1]{\textcolor{gray}{\underline{#1}}}

\begin{document}

\title{VAEmo: Efficient Representation Learning for Visual-Audio Emotion with Knowledge Injection}

\author{Hao Cheng}
\authornote{Both authors contributed equally to this research.}
\email{chenghao@mail.hfut.edu.cn}
\affiliation{%
  \institution{Hefei University of Technology}
  \city{Hefei}
  \country{China}
}
\author{Zhiwei Zhao}
\authornotemark[1]
\email{zhiweizhao@hfut.edu.cn}
\affiliation{%
  \institution{Hefei University of Technology}
  \city{Hefei}
  \country{China}
}
\author{Yichao He}
\email{heyichao@mail.hfut.edu.cn}
\affiliation{%
  \institution{Hefei University of Technology}
  \city{Hefei}
  \country{China}
}
\author{Zhenzhen Hu}
\email{huzhen.ice@gmail.com}
\affiliation{%
  \institution{Hefei University of Technology}
  \city{Hefei}
  \country{China}
}
\author{Jia Li}
\authornote{Corresponding author.}
\email{jiali@hfut.edu.cn}
\affiliation{%
  \institution{Hefei University of Technology}
  \city{Hefei}
  \country{China}
}
\author{Meng Wang}
\email{eric.mengwang@gmail.com}
\affiliation{%
  \institution{Hefei University of Technology}
  \city{Hefei}
  \country{China}
}
\author{Richang Hong}
\email{hongrc.hfut@gmail.com}
\affiliation{%
  \institution{Hefei University of Technology}
  \city{Hefei}
  \country{China}
}

\renewcommand{\shortauthors}{Hao Cheng et al.}

\begin{abstract}
Audiovisual emotion recognition (AVER) aims to infer human emotions from nonverbal visual-audio (VA) cues, offering modality-complementary and language-agnostic advantages. However, AVER remains challenging due to the inherent ambiguity of emotional expressions, cross-modal expressive disparities, and the scarcity of reliably annotated data. Recent self-supervised AVER approaches have introduced strong multimodal representations,  yet they predominantly rely on modality-specific encoders and coarse content-level alignment, limiting fine-grained emotional semantic modeling. To address these issues, we propose VAEmo, an efficient two-stage framework for emotion-centric joint VA representation learning with external knowledge injection. In Stage~1, a unified and lightweight representation network is pre-trained on large-scale speaker-centric VA corpora via masked reconstruction and contrastive objectives, mitigating the modality gap and learning expressive, complementary representations without emotion labels. In Stage~2, multimodal large language models automatically generate detailed affective descriptions according to our well-designed chain-of-thought prompting for only a small subset of VA samples; these rich textual semantics are then injected by aligning their corresponding embeddings with VA representations through dual-path contrastive learning, further bridging the emotion gap. Extensive experiments on multiple downstream AVER benchmarks show that VAEmo achieves state-of-the-art performance with a compact design, highlighting the benefit of unified cross-modal encoding and emotion-aware semantic guidance for efficient, generalizable VA emotion representations. Source code and pre-trained models will be available at \url{https://github.com/MSA-LMC/VAEmo}.
\end{abstract}




\maketitle

\section{Introduction}

\begin{figure}[t]
  \centering
  \includegraphics[width=\linewidth]{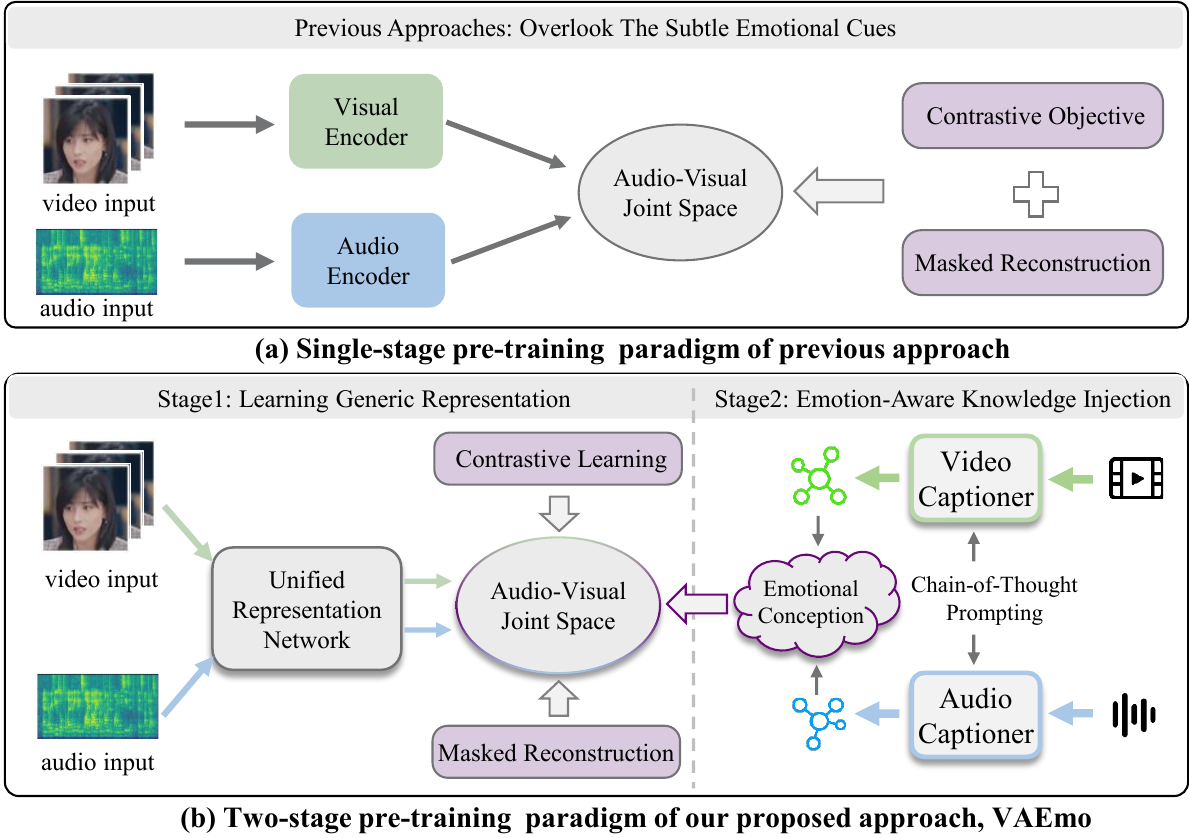}
  \caption{Comparison of pre-training paradigms: (a) prior single-stage approaches focus on learning general audio-visual representations, but lack explicit emotion modeling; (b) our two-stage approach, VAEmo, first builds a joint VA space and then injects external emotion-aware knowledge, enabling fine-grained, emotion-centric learning.}
  \label{fig:motivation}
\end{figure}

Understanding nuanced human emotion is a critical capability for next-generation AI systems that pursue natural and empathetic interactions with users~\cite{liu2024towards,bravo2025systematic,huang2019feeling,lazic2025next,magni2024digital}. Human affect is inherently multimodal, expressed through vocal intonation, facial expressions, body postures, and linguistic content—each modality providing complementary cues for emotion analysis. This study focuses on the task of audio-visual emotion recognition (AVER), which aims to infer human affective states from audio-visual signals in a language-agnostic and modality-complementary manner~\cite{mehrabian2017communication,wu2014survey}.

Emotion recognition is a fine-grained and semantically complex task, where the gap between general content features and emotion-specific features is often more challenging to bridge than the gap between modalities themselves. For instance, a person’s talking face might show sadness in one clip and anger in another, despite consistent identity and modality inputs, the emotional semantics diverge significantly. This indicates the necessity of learning more nuanced and comprehensive emotional representations. Unfortunately, emotional information is subtle, subjective, and difficult to precisely quantify and annotate at scale~\cite{kollias2018aff,9039580,wang2022ferv39k}, hindering conventional supervised approaches and making it challenging to develop powerful multimodal emotion representations~\cite{liu2022mafw,zhang2023transformer,li2023intensity,chen2024static}.

\sloppy
To overcome these limitations in emotion recognition, this field has increasingly turned to self-supervised learning (SSL) techniques~\cite{he2022masked,tong2022videomae,huang2022masked,radford2021learning,hsu2021hubert}. By leveraging large amounts of unlabeled audio-visual data, SSL methods have shown great potential for learning rich multimodal representations without requiring expensive emotion annotations~\cite{gong2022contrastive,huang2023mavil,zellers2021merlot}. However, existing SSL methods, while proficient at capturing general audio-visual correspondences, often overlook subtle emotional cues that require specialized attention. This creates a fundamental mismatch for emotion recognition, as these methods fail to target the extraction of emotion-specific features and fine-grained emotional semantic alignment during pre-training. Consequently, models derived from these self-supervised manner often miss emotional nuances, unable to effectively capture intra-modal emotional cues or align fine-grained emotional semantics across modalities in AVER downstream tasks.

Meanwhile, the emergence of Multimodal Large Language Models (MLLMs)~\cite{xu2024secap} offers promising new directions for addressing these challenges. Although general large models still show moderate performance on representative emotion recognition tasks~\cite{lian2024gpt,xing2024emo}, they encode rich world knowledge and latent commonsense priors that are often neglected by common self-supervised approaches. Incorporating such knowledge during pre-training could significantly benefit emotion representation learning. Prior work like FineCLIPER~\cite{chen2024finecliper} and Emotion-LLaMA~\cite{cheng2024emotion} has begun exploring large models by generating captions to enhance model learning. However, FineCLIPER focuses solely on the visual modality, ignoring multimodal information, and operates during fine-tuning stage rather than pre-training stage. Its inference requires captions for additional supervision, limiting rapid transfer to downstream tasks. Similarly, Emotion-LLaMA only extracts peak frames from videos for facial action unit analysis, making its performance heavily reliant on the accuracy of peak frame selection, which is often limited in in-the-wild environments.

In response to these challenges, we introduce VAEmo, a novel framework specifically designed for audio-visual emotion representation learning with a two-stage training paradigm. Inspired by recent works on unified encoder designs for multimodal learning~\cite{srivastava2024omnivec2, bachmann2022multimae, zhang2023meta}, our VAEmo adopts a an efficient unified encoder that processes both audio and visual inputs, significantly reducing model 
parameters while maintaining excellent recognition performance . Unlike previous approaches that typically use separate encoders for each modality (dual-tower models), our unified encoder enables tighter integration of multimodal features, facilitating more efficient and effective representation learning. As illustrated in Fig.~\ref{fig:motivation}, our approach first establishes foundational cross-modal correspondences through contrastive and masked reconstruction learning (Stage~1) and then injects emotion-specific knowledge using MLLMs (Stage~2).

Our knowledge injection pipeline leverages specialized MLLMs to generate emotion-focused descriptions from both modalities, creating a rich semantic bridge between raw signals and emotional concepts. Through a dual-path contrastive learning approach that maintains separate audio-text and video-text learning paths while optimizing the same representation network, we preserve modality-specific characteristics while enriching emotional semantics. Our carefully designed two-stage training paradigm effectively addresses the gap between general audio-visual representation learning and emotion-specific feature extraction. 
By first establishing strong cross-modal
representations in Stage~1 before injecting emotion-specific knowledge in Stage~2, we create a more stable learning trajectory. 
Extensive experiments demonstrate that VAEmo consistently outperforms previous state-of-the-art approaches while using significantly fewer parameters.
In summary, our work makes the following contributions:
\begin{itemize}
  \item A novel two-stage training paradigm that progressively transitions from generic audio-visual representations into emotion-centric representations through contrastive learning, masked prediction, and knowledge injection.
  \item An innovative knowledge injection method leveraging MLLMs to generate emotion-focused descriptions, creating a rich semantic bridge between raw signals and emotional concepts through a dual-path contrastive learning approach.
  \item Comprehensive experimental validation revealing that for emotion-specific tasks, architectural simplicity coupled with targeted knowledge injection yields superior results compared to more complex models.
\end{itemize}

\section{Related Work}
\label{sec:relatedwork}
\begin{figure*}[!t]
    \centering
    \includegraphics[width=\linewidth]{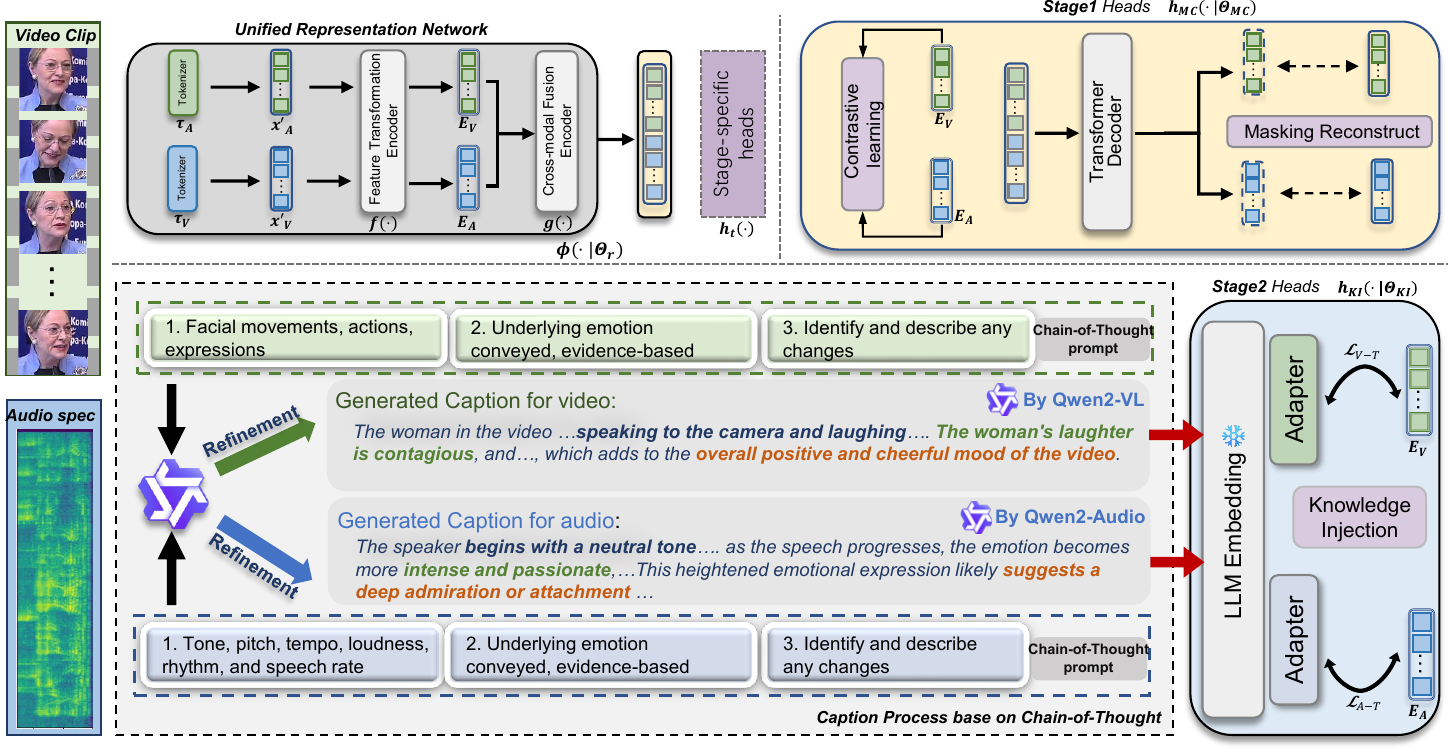}
    \caption{The architecture of our VAEmo framework. VAEmo adopts a two-stage training paradigm for audio-visual emotion recognition. Raw inputs are first encoded by the representation network, which comprises three main components: modality-specific tokenizers, a shared-parameter feature transformation encoder, and a cross-modal fusion encoder. In Stage~1 (upper), masked autoencoding and contrastive learning objectives are applied to learn aligned audio-visual representations. Stage~2 (below) builds upon this foundation by injecting external emotional knowledge: captions generated via a Chain-of-Thought (CoT)-based module are incorporated through dual-path contrastive learning.}
    
    \label{fig:model}

\end{figure*}

\noindent \textbf{Audio-Visual Emotion Recognition.} Audio-visual emotion recognition (AVER) integrates speech and facial expression analysis, providing complementary multi-modal perspectives on emotion understanding~\cite{mehrabian2017communication,wu2014survey}. Early approaches relied on supervised learning with hand-crafted features, such as eGeMAPS~\cite{eyben2015geneva} for audio and VGGFace~\cite{cao2018vggface2} for visual features. With the advancement of deep learning, end-to-end approaches became prevalent. Liu et al.~\cite{liu2022mafw} proposed T-ESFL, which employs collaborative multi-task learning to improve AVER performance. Zhang et al.~\cite{zhang2023transformer} developed T-MEP, introducing attention mechanisms to enhance cross-modal interaction. Recently, self-supervised approaches have made significant breakthroughs, such as AW-HuBERT~\cite{tran2023saaml}  leveraged audio-visual self-supervised pre-training to enhance scene adaptation capabilities, and  HiCMAE~\cite{sun2024hicmae} learned expressive representations through hierarchical contrastive masked autoencoding. Additionally, large models have been applied to emotion analysis.
Chen et al.~\cite{chen2024finecliper} introduced FineCLIPER, which utilizes LLMs to generate emotional annotations, and Zhao et al.~\cite{zhao2023prompting} proposed DFER-CLIP, which explores visual prompt tuning for emotion recognition.
In contrast to these methods, our VAEmo adopts a compact single-stream architecture and two-stage training paradigm, significantly enhancing emotional representation capacity while maintaining high efficiency.

\noindent \textbf{Self-Supervised Learning and Knowledge Injection.} Self-supervised learning (SSL) has become the dominant paradigm for representation learning. In the visual domain, MAE~\cite{he2022masked}  employs high-ratio masked reconstruction pre-training, while Tong et al.~\cite{tong2022videomae} extended this to the spatiotemporal domain. For audio, Hsu et al.'s~\cite{hsu2021hubert} HuBERT and Chen et al.'s~\cite{chen2022wavlm} WavLM utilize hidden unit prediction to establish powerful speech representations. In multi-modal learning, Gong et al.~\cite{gong2022contrastive} combined contrastive learning with MAE, while MAViL~\cite{huang2023mavil}  explored more on contrastive object. 
Despite these advances, existing SSL approaches primarily focus on general-purpose representation learning, often overlooking the injection of domain-specific knowledge such as emotion semantics.
For knowledge injection, with the emergence of large models, researchers have begun exploring LLMs as external knowledge sources.
LLM2CLIP~\cite{wu2024llm2clip} injects knowledge from large models into CLIP~\cite{radford2021learning} models to enhance the performance of visual-language models. MATE~\cite{jang2024mate} uses LoRA-tuned~\cite{hu2022lora} large language models to enhance the performance of CLIP visual encoders.
Different from these works, our method uniquely uses multimodal LLMs (Qwen2-VL~\cite{wang2024qwen2} and Qwen2-Audio~\cite{chu2024qwen2}) to generate emotion-focused descriptions, incorporating emotional knowledge through dual-path contrastive learning that effectively bridges semantic gap between general and emotion-specific representations.

\section{Approach}
\label{sec:method}

\textbf{Problem Formulation.} We focus on learning emotion representations from the vast amount of audio and video data available on the Internet. With audio and video modalities indexed by $m \in\{A, V\}$, each modality provides complementary emotional cues through different channels—speech patterns (e.g., tone, prosody) for audio and facial dynamics (e.g., expressions, micro-movements) for video. We propose a simple yet powerful model architecture coupled with a two-stage training framework to capture these emotional representations effectively.

\noindent
\textbf{Overview.} As shown in Fig.~\ref{fig:model}, our method follows a two-stage training paradigm. Raw inputs are encoded through our representation network, which consists of three key components: modality-specific tokenizers, a feature transformation encoder, and a cross-modal fusion encoder. Based on different training stages, the encoded audio and visual features are input to stage-specific task heads. In Stage~1, this includes a combination of masked reconstruction and contrastive learning objectives, facilitating the establishment of cross-modal representations. In Stage~2, this becomes our knowledge injection framework, where captions generated by our carefully designed Chain of Thought (CoT)-based caption generation mechanism serve as emotional knowledge injected into representation network. The following sections provide detailed descriptions of VAEmo's architecture and two-stage training paradigm.

\subsection{Network Architecture}
VAEmo architecture consists of four main components as shown in Fig.~\ref{fig:model}: (1) \textit{Modality-specific tokenizers $\tau_m(\cdot)$}: convert raw audio and visual inputs into token sequences with consistent dimensions; (2) \textit{A shared-parameter feature transformation encoder $f(\cdot)$}: 
encode each modality with multiple Transformer layers using shared parameters.
(3) \textit{A cross-modal fusion encoder $g(\cdot)$}: integrates information by concatenating outputs from $f(\cdot)$ across modalities, followed by Transformer layers
with shared parameters; (4) \textit{Stage-specific heads $h_t(\cdot)$}: vary depending on the training stage $t$.

Both the feature transformation encoder and fusion encoder are implemented using a Vision Transformer (ViT)-based architecture. This intentionally simplified design offers two significant advantages: (1) it allows us to focus research attention on our novel two-stage training methodology rather than architectural innovations, providing clearer insights into the effectiveness of our knowledge injection approach; and (2) the unified architecture with shared parameters across modalities enables tighter integration of multimodal features in our encoder, while reducing the model's parameter count compared to traditional dual-stream methods commonly used in audio-visual representation learning.

\noindent
\textbf{Data Processing:} Given unlabeled video data with accompanying audio tracks $x \in \mathcal{D}$, we demultiplex and pre-process the data into a video frame sequence $x_V\in{\mathbb{R}^{T_V\times H\times W \times 3}}$ and an audio log-mel spectrogram $x_A\in{\mathbb{R}^{T_a \times F}}$ for the respective modalities. These raw inputs are transformed into token sequences $x^{'}_{V}\in{\mathbb{R}^{N_v\times D}}$ and $x^{'}_{A} \in \mathbb{R}^{N_a\times D}$ using modality-specific tokenizers as shown in Fig.~\ref{fig:model}.
For audio, we treat mel-spectrograms as single-channel images and apply non-overlapping 2D convolutions (patch size $16 \times 16$) to tokenization. For video, we employ 3D convolutions to extract spatiotemporal "tubes" from frame sequences. This yields $N_a = \frac{T_a}{16} \times \frac{F}{16}$ audio tokens and $N_v = \frac{T_V}{2} \times \frac{H}{16} \times \frac{W}{16}$ video tokens. This tokenization process ensures consistent output dimensions across modalities. The tokenizers thus produce dimension-aligned sequences $x'_m \in \mathbb{R}^{N_m \times D}$, where $D$ is the dimension of the embedding.

Our model architecture consists of a unified representation network $\phi(x|\Theta_r)$ and stage-specific task heads $h_t(z|\Theta_t)$.
\begin{equation}
  \phi(x|\Theta_r) = g(f(\tau_A(x_A)) \oplus f(\tau_V(x_V))) = g(f(x'_A) \oplus f(x'_V))
\end{equation}
where $\tau_A$ and $\tau_V$ represent the modality-specific tokenizers, $\Theta_r$ represents the representation network parameters, and $\oplus$ denotes token-wise concatenation of feature sequences. The data flow for a specific training stage $t$ can be expressed as:
\begin{equation}
 \psi_{t}(x|\Theta_r,\Theta_{t}) = h_t(\phi(x|\Theta_r); x, E_V,E_A|\Theta_{t})
\end{equation}


The overall learning process involves optimizing both representation parameters $\Theta_r$ and Stage-specific parameters $\Theta_t$:
\begin{equation}
\Theta^* = \min_{\Theta} \mathbb{E}_{x \sim \mathcal{D}}[\mathcal{L}_t ( \psi_t(x|\Theta) )]
\end{equation}
where $\Theta = \{\Theta_r, \Theta_t\}$ and $\mathcal{L}_t$ denotes the loss function for training stage $t$. While the task heads $h_t$ change across different stages, they all optimize the same underlying representation network $\phi$, allowing us to effectively transfer knowledge across stages.

\noindent
\textbf{Stage-Specific Task Heads:} Our framework employs three distinct task heads across training stages:

\textbf{Stage~1:} A dual-purpose head ($h_{MC}$) supporting both masked reconstruction and contrastive learning objectives. The reconstruction component decodes masked tokens while the contrastive component aligns audio-visual representation.

\textbf{Stage~2:} A knowledge injection head ($h_{KI}$) that facilitates dual-path contrastive learning between audio-text and video-text pairs using MLLM-generated descriptions.

\textbf{Fine-tuning:} A simple classification head ($h_{cls}$) implemented as a single-layer MLP that projects the fused representations to emotion categories or dimensional scores.

\subsection{Training Paradigm}

\textbf{Stage~1: Contrastive Learning and Masked Reconstruction.}
For Stage~1 pre-training, we employ a dual strategy combining contrastive learning and masked self-supervised learning, an approach proven effective in prior works like CAV-MAE~\cite{gong2022contrastive} and MAViL~\cite{huang2023mavil}. 
Our contrastive objective employs the standard InfoNCE loss with temperature scaling to align the global representations of paired video and audio. 
For each video–audio pair, the global representations for video and audio are obtained by token-wise mean pooling over the output feature sequences  $E_V$ and $E_A$, respectively.
Simultaneously, we perform masked self-supervised learning by randomly masking tokens (80\% for audio, 90\% for video) and tasking the model to reconstruct the original data from these partially masked inputs. The reconstruction process involves processing masked inputs through our encoders, fusing modality-specific features, and computing the mean squared error loss between original and reconstructed data. This combined approach offers complementary benefits: contrastive learning captures high-level cross-modal correspondences, while reconstruction preserves fine-grained details within each modality.
After Stage~1 pre-training, we discard the reconstruction decoder and retain only the representation network $\phi$ for Stage~2, allowing the model to first learn generalizable speech-face representations before incorporating emotional knowledge.

\begin{figure}[!t]
  \centering
  \includegraphics[width=\linewidth]{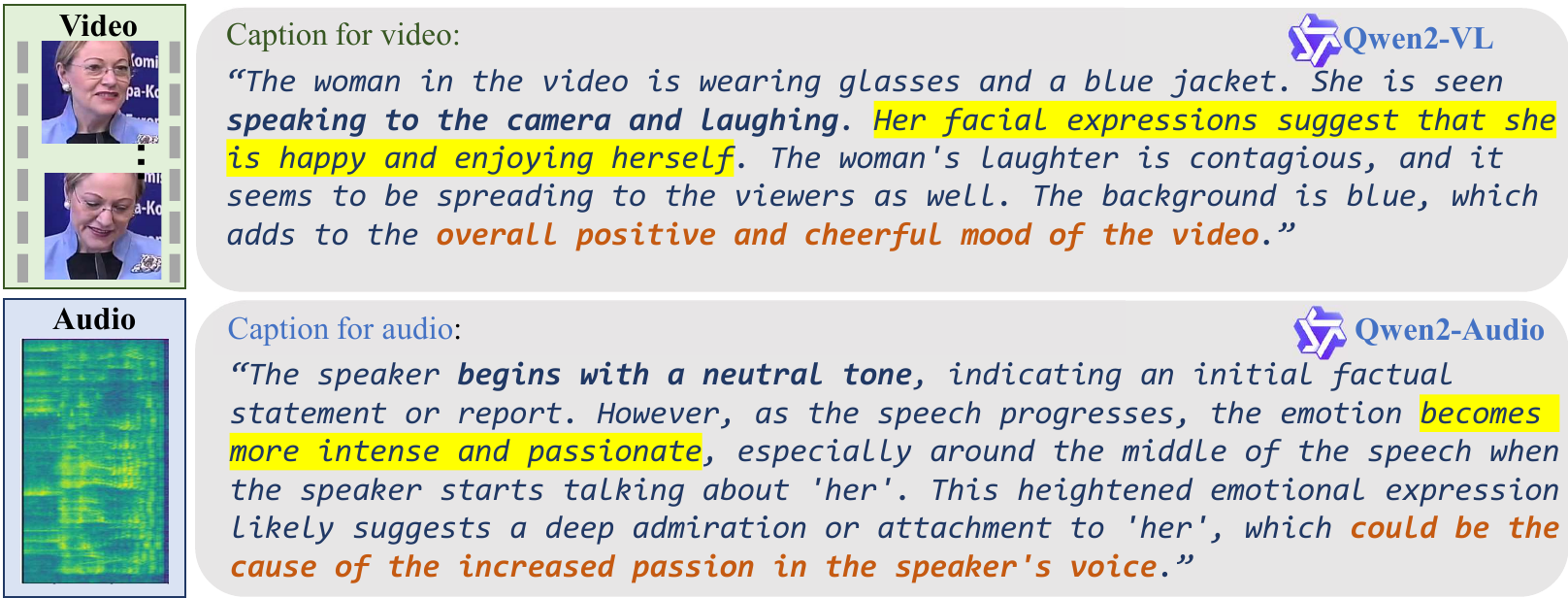}
  \caption{Examples of emotion-focused captions generated by our Chain-of-Thought (CoT) prompting strategy.}
  \label{fig:caption}

\end{figure}

\noindent
\textbf{Stage~2: Knowledge Injection with MLLMs.}
In the second stage, we further enhance our model's emotion representation capabilities based on the stage~1 parameters 
through knowledge injection. We develop a fully automated knowledge extraction and injection pipeline leveraging Multimodal Large Language Models (MLLMs). Using modality-specific MLLMs—Qwen2-VL for video and Qwen2-Audio for audio, we generate rich emotional descriptions through a carefully designed Chain-of-Thought (CoT) prompting strategy. 

Our CoT process first guides MLLMs to describe observable physical characteristics—facial action units, overall posture, and body movements for video, or acoustic properties for audio. Based on these objective features, MLLMs then assess emotional states, including intensity (arousal) and valence, and finally analyze overall emotional changes throughout the sequence. To ensure that the generated captions are both factual and emotionally relevant, we introduce a refinement mechanism. First, we explicitly instruct the MLLMs to ground all inferences in observable features. Then, we conduct multiple inference passes per input and apply a majority voting strategy to filter out inconsistent or hallucinated outputs. We employ a secondary LLM (GPT-4o-mini~\cite{hurst2024gpt}) to filter the results by discarding emotion-irrelevant content.This refinement pipeline encourages the generation of expressive and specialized affective vocabulary (e.g., “assertiveness,” “amusement,” “demeanor”), enabling the model to capture subtle and compound emotional states. As illustrated in Fig.~\ref{fig:caption}, these high-quality descriptions serve as valuable supervision signals during training.

Since the generated captions are much longer than CLIP~\cite{radford2021learning}'s 77-token limit (in our implementation, tokens counted using GPT-2~\cite{radford2019language} tokenizer can reach up to 100) and contain specialized emotional vocabulary absent from CLIP's training data, we employ LLM-based embeddings for text representation. Specifically, we use the NV-Embed-v2~\cite{lee2024nv} model, which adapts a decoder-only LLM as an embedding generator by employing specialized latent attention layers and removing causal attention masks. Our ablation experiments demonstrate that LLM embeddings significantly outperform both CLIP text embeddings and BERT embeddings for our emotion recognition tasks.
After encoding the captions into text embeddings, we further process each embedding through a modality-specific adapter module, resulting in $C_{A}$ and $C_{V}$ for audio and video, respectively. The adapter design follows the implementation in LLM2CLIP~\cite{wu2024llm2clip}. We then implement a dual-path contrastive learning approach (Fig.~\ref{fig:model}), which replaces the traditional audio-visual contrastive objective from Stage~1. Specifically, we construct two parallel contrastive objectives: one between audio caption embedding and audio global representation, and the other between video caption embedding and video global representation.

\begin{equation}
  \mathcal{L}_{m-T} = -\log \frac{\exp(sim(z_m, C_m)/\tau)}{\sum_{j=1}^{N} \exp(sim(z_m, C_m^j)/\tau)}
\end{equation}
\noindent where $ m \in \{A, V\} $ represents either audio or video. The total loss for Stage~2  is a weighted combination of these components:
\begin{equation}
\mathcal{L}_{Stage2} = \alpha\mathcal{L}_{A-T} + \beta\mathcal{L}_{V-T}
\label{eq:equation_5}
\end{equation}
A crucial aspect of this dual-path approach is that both objectives simultaneously optimize the same encoder through different gradient paths. $z_A$ and $z_V$ are the token-wise mean-pooled embedding of the feature transformation encoder output  for audio and video, respectively. The gradients from the two contrastive losses flow back to update the same set of parameters. This creates a rich training signal that enhances the encoder's ability to aware emotional knowledge for both modalities.


After completing both training stages, we extract the encoder and fine-tune it on downstream tasks to leverage the emotion-rich representations for specific applications. The sequential nature of our training paradigm—from general audio-visual correspondences to emotion-specific knowledge—creates a robust foundation for visual-audio emotion recognition.

\section{Experiments}
\label{sec:experiments}

In this section, we evaluate the effectiveness of VAEmo on a suite of audio-visual emotion recognition (AVER) tasks. Our experiments cover both categorical and dimensional emotion recognition paradigms across diverse datasets, spanning from controlled lab environments to unconstrained in-the-wild scenarios.

\subsection{Experimental Setup}

\begin{table}[t]
  \centering
  \caption{Datasets overview in our experiments. "Wild" indicates in-the-wild data, "Lab" indicates laboratory-controlled settings. UAR: Unweighted Average Recall, WAR: Weighted Average Recall, PCC: Pearson Correlation Coefficient.}
  \label{tab:datasets}
  \resizebox{\columnwidth}{!}{
  \begin{tabular}{lccccc}
    \toprule
    Dataset & Setting & Task & \#Classes & Samples & Metric \\
    \midrule
    VoxCeleb2~\cite{chung2018voxceleb2} & - & Pre-training & - & 1,360,531 & - \\
    \midrule
    MAFW~\cite{liu2022mafw} & Wild & Categorical & 11 & 9,172 & UAR, WAR \\
    DFEW~\cite{jiang2020dfew} & Wild & Categorical & 7 & 11,697 & UAR, WAR \\
    CREMA-D~\cite{cao2014crema} & Lab & Categorical & 6 & 7,442 & UAR, WAR \\
    MSP-IMPROV~\cite{busso2016msp} & Lab & Categorical & 4 & 7,798 & UAR, WAR \\
    \midrule
    Werewolf-XL~\cite{zhang2021werewolf} & Lab & Dimensional & 3 & 8,542 & PCC \\
    \bottomrule
  \end{tabular}
  }
\end{table}

\textbf{Datasets.} Tab.~\ref{tab:datasets} provides a comprehensive overview of all datasets used in our experiments; all used datasets are video data with accompanying audio tracks. For pre-training (including both Stage~1 and Stage~2), we use VoxCeleb2, which contains over 1 million utterances from 6,112 celebrities extracted from YouTube videos. This dataset provides a rich source of speech-face pairs for learning audio-visual representations. For downstream evaluation, we select representative datasets covering different aspects of emotion recognition: MAFW and DFEW for in-the-wild categorical emotion recognition, which present challenging scenarios with natural variations in lighting, pose, and acoustic conditions; CREMA-D and MSP-IMPROV for lab-controlled categorical emotion recognition with high-quality recordings under controlled settings; and Werewolf-XL for dimensional emotion recognition, which annotates emotions along continuous dimensions including valence, arousal, and dominance.

\noindent \textbf{Evaluation Metrics.} For categorical emotion recognition, we report both Unweighted Average Recall (UAR) and Weighted Average Recall (WAR) to address potential class imbalance issues. For dimensional emotion recognition, we use the Pearson Correlation Coefficient (PCC) following standard practice. For those datasets using cross-validation, we combine the predictions and labels from all folds to calculate the final UAR and WAR, the same as the evaluation in~\cite{sun2024hicmae}.

\subsection{Implementation Details}

\textbf{Preprocessing.} For video preprocessing, we employ different strategies during pre-training and fine-tuning. During pre-training, we follow~\cite{sun2023mae} to crop each frame to $160 \times 160$ pixels and sample 16 frames to form video clips with dimensions $16 \times 160 \times 160 \times 3$ (\textit{frames $\times$ height $\times$ width $\times$ channels}), which is computationally efficient for large-scale processing. For fine-tuning on downstream tasks, we employ InsightFace~\cite{insightface} to detect and crop facial regions more precisely from each frame. The facial crops are resized to $160 \times 160$ pixels. For audio, we extract 128-dimensional mel-spectrograms with a 25~ms window size and 10~ms shift, which are then converted to single-channel "images" for processing, following~\cite{sun2024hicmae}.

\noindent \textbf{Model Configuration.} Our VAEmo framework employs a ViT-Small architecture. The model maintains a consistent embedding dimension of 512 across all components with 8 attention heads. Our streamlined architecture consists of a feature transformation encoder $f(\cdot)$ which is 10-layer transformer followed by a fusion encoder $g(\cdot)$ which is 2-layer transformer, resulting in a total parameter count of 39M—significantly smaller than comparable models like HiCMAE-B (81M) and AV-HuBERT variants (103M).

\noindent \textbf{Training Details.} Our training proceeds in two distinct stages. For Stage~1 Pre-training, we train the model on the full VoxCeleb2 dataset using 8 NVIDIA A800 GPUs with a batch size of 170 per GPU. We use the AdamW optimizer with a base learning rate of 1.2e-3 and a cosine learning rate schedule. This stage involves 100 epochs of training, requiring approximately 48 hours to complete. For Stage~2 Knowledge Injection, to incorporate emotional knowledge, we randomly sample 10\% of the VoxCeleb2 dataset (approximately 100K samples) and generate emotion-focused captions using our MLLM pipeline. To prevent catastrophic forgetting, we employ LayerNorm tuning~\cite{zhao2023tuning} where only LayerNorm parameters are updated while keeping other weights frozen. This stage involves 10 epochs of training on the captioned subset.

\subsection{Comparison with State-of-the-Art Methods }
\begin{table}[!t]
  \centering
  \caption{Performance comparison on MAFW dataset (11-class). The best results are highlighted in \textbf{Bold}, and the second-best \underline{Underlined}.}
  \label{tab:mafw}
  \small
  \begin{tabular}{lccccc}
      \toprule
      Method & SSL & Modality & \#Params(M) & UAR & WAR \\
      \midrule
      \multicolumn{6}{l}{\textit{Audio-only approaches:}} \\
      \midrule
      HuBERT~\cite{hsu2021hubert} & \checkmark & A & 95 & 25.00 & 32.60 \\
      WavLM-Plus~\cite{chen2022wavlm} & \checkmark & A & 95 & 26.33 & 34.07 \\
      \midrule
      \multicolumn{6}{l}{\textit{Visual-only approaches:}} \\
      \midrule
      DFER-CLIP~\cite{zhao2023prompting} & \checkmark & V & 153 & 39.89 & 52.55 \\
      SVFAP~\cite{sun2024svfap} & \checkmark & V & 78 & 41.19 & 54.28 \\
      MAE-DFER~\cite{sun2023mae} & \checkmark & V & 85 & 41.62 & 54.31 \\
      S4D~\cite{chen2025staticdynamicdeeperunderstanding} & \checkmark & V & 101 & 43.72 & 58.44 \\
      \midrule
      \multicolumn{6}{l}{\textit{Multi-modal approaches:}} \\
      \midrule
      T-ESFL~\cite{liu2022mafw} & $\times$ & A+V & - & 33.35 & 48.70 \\
      T-MEP~\cite{zhang2023transformer} & $\times$ & A+V & 61 & 37.17 & 51.15 \\
      MMA-DFER~\cite{chumachenko2024mma} & \checkmark & A+V & - & 44.25 & \underline{58.45} \\
      HiCMAE-S~\cite{sun2024hicmae} & \checkmark & A+V & 46 & 41.66 & 54.45 \\
      HiCMAE-B~\cite{sun2024hicmae} & \checkmark & A+V & 81 & 42.65 & 56.17 \\
      \rowcolor{teal!20}
      VAEmo (ours) & \checkmark & A+V & \textbf{39} & \textbf{45.67} & \textbf{58.91} \\
      \midrule
      \textcolor{gray}{FineCLIPER~\cite{chen2024finecliper}} & \textcolor{gray}{\checkmark} & \textcolor{gray}{T+V} & \textcolor{gray}{N/A} & \grayunderline{45.01} & \textcolor{gray}{56.91} \\
      \bottomrule
  \end{tabular}
\end{table}

\begin{table}[t]\small
  \centering
  \caption{Performance comparison on DFEW dataset. Approaches grouped by pre-training strategy: (A) approaches using external pre-trained weights or additional labels, (B) approaches trained from scratch using only VoxCeleb2, enabling fair comparison with our approach.}
  \label{tab:dfew}
  \begin{tabular}{lccccc}
      \toprule
      Method & SSL & Mod. & \#Params(M) & UAR & WAR \\
      \midrule
      \multicolumn{6}{l}{\textit{(A) Approaches using external pre-trained weights or additional labels:}} \\
      \midrule
      S2D~\cite{chen2024static} & \checkmark & V & N/A & \underline{65.45} & \underline{74.81} \\
      S4D~\cite{chen2025staticdynamicdeeperunderstanding} & \checkmark & V & 101 & \textbf{66.80} & \textbf{76.68} \\
      \textcolor{gray}{FineCLIPER~\cite{chen2024finecliper}} & \textcolor{gray}{\checkmark} & \textcolor{gray}{T+V} & \textcolor{gray}{N/A} & \textcolor{gray}{65.98} & \textcolor{gray}{76.21} \\
      \midrule
      \multicolumn{6}{l}{\textit{(B) Approaches trained from scratch on VoxCeleb2 only:}} \\
      \midrule
      MAE-DFER~\cite{sun2023mae} & \checkmark & V & 85 & 63.41 & 74.43 \\
      HiCMAE-S~\cite{sun2024hicmae} & \checkmark & A+V & 46 & 63.05 & 74.33\\
      HiCMAE-B~\cite{sun2024hicmae} & \checkmark & A+V & 81 & \underline{63.76} & \underline{75.01} \\
      \rowcolor{teal!20}
      VAEmo (ours) & \checkmark & A+V & \textbf{39} & \textbf{64.02} & \textbf{75.78} \\
      \midrule
      AMH~\cite{yoon2020attentive} & $\times$ & A+V & - & 54.48 & 66.51 \\
      \bottomrule
  \end{tabular}
\end{table}

\begin{table}[t]\small
  \centering
  \caption{Performance comparison on the lab-controlled categorical emotion recognition dataset CREMA-D.}
  \label{tab:cremad}
  \begin{tabular}{lccccc}
      \toprule
      Method & SSL & Mod. & \#Params(M) & UAR & WAR \\
      \midrule
      HuBERT~\cite{hsu2021hubert} & \checkmark & A & 95 & 72.72 & 72.57 \\
      WavLM-Plus~\cite{chen2022wavlm} & \checkmark & A & 95 & 73.34 & 73.39 \\
      SVFAP~\cite{sun2024svfap} & \checkmark & V & 78 & 77.31 & 77.37 \\
      MAE-DFER~\cite{sun2023mae} & \checkmark & V & 85 & 77.33 & 77.38 \\
      \midrule
      VQ-MAE-AV~\cite{hazmoune2024using} & \checkmark & A+V & 30 & - & 80.40 \\
      HiCMAE-S~\cite{sun2024hicmae} & \checkmark & A+V & 46 & 84.46 & 84.38 \\
      HiCMAE-B~\cite{sun2024hicmae} & \checkmark & A+V & 81 & 84.91 & 84.89 \\
      \rowcolor{teal!20}
      VAEmo (ours) & \checkmark & A+V & \textbf{39} & \textbf{85.71} & \textbf{85.68} \\
      \bottomrule
  \end{tabular}
\end{table}

\begin{table}[t]\small
  \centering
  \caption{Performance comparison on the lab-controlled categorical emotion recognition dataset MSP-Improv.}
  \label{tab:mspimprov}
  \begin{tabular}{lccccc}
      \toprule
      Method & SSL & Mod. & \#Params(M) & UAR & WAR \\
      \midrule
      FAV-HuBERT~\cite{tran2023saaml} & \checkmark & A+V & 103 & 61.05 & 68.35 \\
      TAPT-HuBERT~\cite{tran2023saaml} & \checkmark & A+V & 103 & 63.95 & 70.46 \\
      AW-HuBERT~\cite{tran2023saaml} & \checkmark & A+V & 103 & 65.72 & 71.80 \\
      HiCMAE-S~\cite{sun2024hicmae} & \checkmark & A+V & 46 & 63.90 & 74.35 \\
      HiCMAE-B~\cite{sun2024hicmae} & \checkmark & A+V & 81 & 65.78 & 74.95 \\
      \rowcolor{teal!20}
      VAEmo (ours) & \checkmark & A+V & \textbf{39} & \textbf{66.12} & \textbf{76.79} \\
      \bottomrule
  \end{tabular}
\end{table}

\begin{table}[t]\footnotesize
  \centering
  \caption{Performance comparison on the dimensional emotion recognition dataset Werewolf-XL.}
  \label{tab:werewolf}
  \begin{tabular}{lcccccc}
      \toprule
      Method & SSL & Mod. & \#Params(M) & Arousal & Valence & Dominance \\
      \midrule
      eGeMAPS~\cite{eyben2015geneva} & $\times$ & A & - & 23.45 & 8.08 & 31.15 \\
      VGGFace~\cite{cao2018vggface2} & $\times$ & V & - & 7.24 & 62.96 & 14.30 \\
      SVFAP~\cite{sun2024svfap} & \checkmark & V & 78 & 23.51 & 67.11 & 34.61 \\
      HiCMAE~\cite{sun2024hicmae} & \checkmark & A+V & 81 & 33.74 & 69.23 & 40.66 \\
      \rowcolor{teal!20}
      VAEmo (ours) & \checkmark & A+V & \textbf{39} & \textbf{39.80} & \textbf{69.72} & \textbf{52.87} \\
      \bottomrule
  \end{tabular}
\end{table}

Tabs.~\ref{tab:mafw}-\ref{tab:werewolf} present comprehensive results comparing our VAEmo framework with state-of-the-art approaches across five diverse emotion recognition benchmarks. Our model demonstrates superior performance while maintaining significantly fewer parameters (39M) than previous approaches (81 M- 103 M). Note that for approaches marked with N/A in parameter count (S2D and FineCLIPER), only fine-tunable parameters (9M and 20M, respectively) were reported in the original papers without specifying the total parameter count. Based on their use of ViT-base as the backbone, we estimate their total parameter count to exceed 86 M.

As shown in Tab.~\ref{tab:mafw}, on the challenging in-the-wild categorical emotion recognition dataset MAFW, our approach achieves 45.67\% UAR and 58.91\% WAR, outperforming the previous best method (MMA-DFER~\cite{chumachenko2024mma}), which achieved 44.25\% UAR and 58.45\% WAR. On the DFEW dataset, while S4D~\cite{chen2025staticdynamicdeeperunderstanding} achieves higher UAR (66.80\%), our method reaches 75.78\% WAR, exceeding HiCMAE-B~\cite{sun2024hicmae} (75.01\% WAR) with only 39M parameters compared to HiCMAE-B's 81M and S4D's 101M parameters. It is worth noting that S4D leverages additional static facial expression datasets for pre-training, while FineCLIPER incorporates extra modalities, including facial landmarks, and S2D similarly utilizes external data resources. Meanwhile, among all approaches that use only VoxCeleb2 for pre-training (as shown in Tab.~\ref{tab:dfew}, group B), VAEmo consistently achieves the best performance while requiring significantly fewer parameters, highlighting the effectiveness of our two-stage training paradigm and knowledge injection approach.

On the lab-controlled categorical emotion recognition dataset CREMA-D, our method achieves 85.71\% UAR and 85.68\% WAR, setting a new benchmark that surpasses larger models like HiCMAE-B~\cite{sun2024hicmae} (84.91\% UAR, 84.89\% WAR). On MSP-IMPROV, we obtain 66.12\% UAR and 76.79\% WAR, outperforming the previous best approach AW-HuBERT~\cite{tran2023saaml} (65.72\% UAR, 71.80\% WAR) by +0.40\% UAR and +4.99\% WAR while using 62\% fewer parameters. These results demonstrate our model's exceptional capability in controlled recording environments.
Our method also excels at continuous emotion recognition on the Werewolf-XL dataset, achieving 39.80\%, 69.72\%, and 52.87\% PCC for arousal, valence, and dominance dimensions, respectively. The improvements compared to HiCMAE~\cite{sun2024hicmae} are particularly substantial for arousal (+6.06\%) and dominance (+12.21\%), indicating our approach's effectiveness at capturing subtle emotional intensity variations.
These results collectively demonstrate that architectural simplicity coupled with our two-stage knowledge injection paradigm creates a powerful foundation for emotion recognition across diverse settings.

\begin{figure}[!t]
  \centering
  \includegraphics[width=\linewidth]{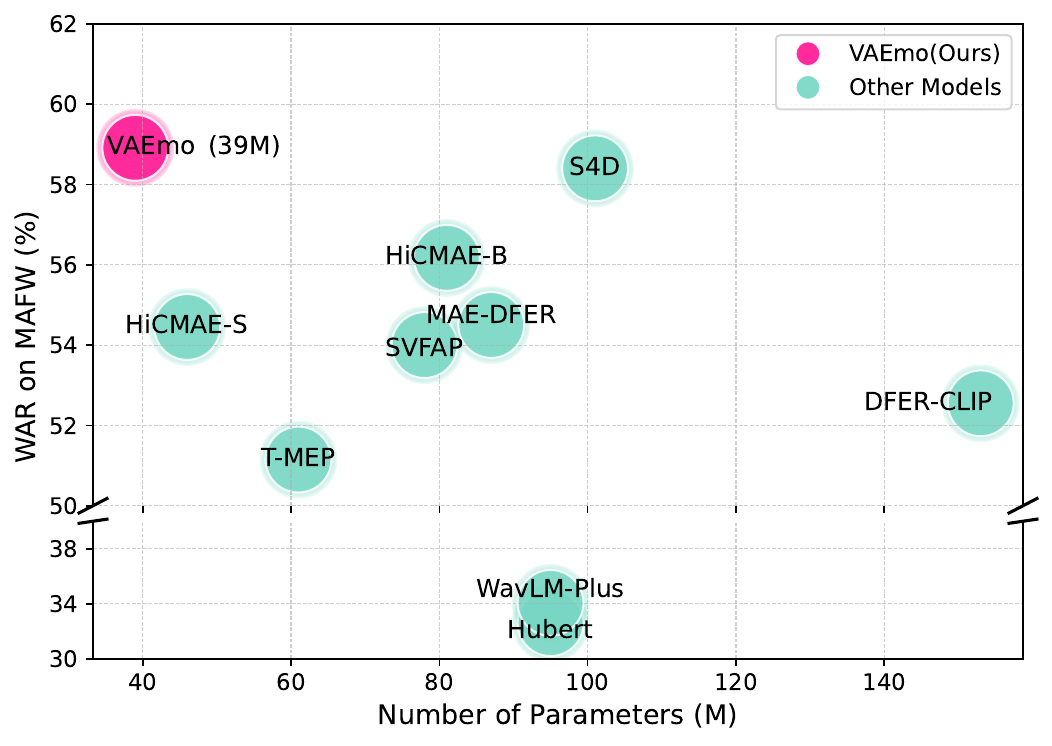}
  \caption{Comparison of the model performance and parameter count on the MAFW dataset. Compared to other state-of-the-art approaches, our proposed VAEmo achieves better performance with fewer parameters.}
  \label{fig:mafw_img}
\end{figure}

Furthermore, as shown in Fig.~\ref{fig:mafw_img}, our proposed model, VAEmo (39M parameters), outperforms other state-of-the-art approaches on the MAFW dataset with a significantly lower parameter count.
Specifically, VAEmo achieves a WAR of 58.91\%, while models such as HiCMAE-B (81M parameters) and S4D (101M parameters) perform at 56.17\% and 58.44\%, respectively. 

\subsection{Ablation Studies}
\label{sec:ablation}


\begin{figure}[!t]
  \centering
  \includegraphics[width=1\linewidth]{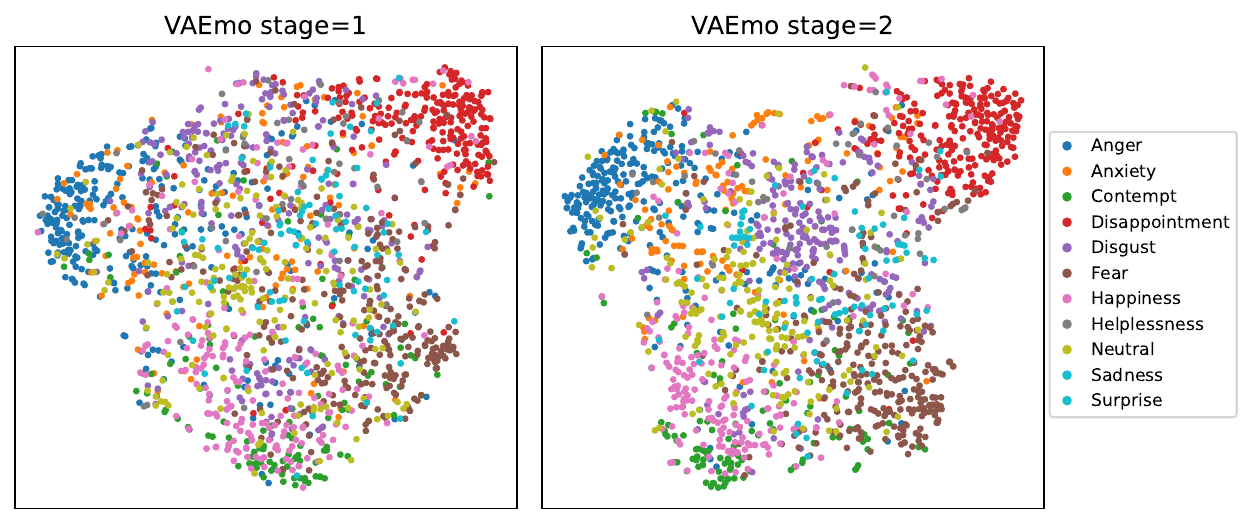}
  \caption{T-SNE visualization of representations from the MAFW dataset comparing Stage~1 and Stage~2 of our VAEmo.}
  \label{fig:mafw_tsne}
  \Description{A bar chart comparing the performance metrics (UAR and WAR) between Stage~1 and Stage~2 of the VAEmo framework on the first fold of the MAFW dataset, showing consistent improvements after knowledge injection.}
\end{figure}

\noindent
\textbf{Effectiveness of the Proposed Two-stage Training.}
\begin{table}[t]
  \centering
  \caption{Ablation study on Two-Stage Training across MAFW, MSP-IMPROV and Werewolf-XL datasets.}
  \label{tab:ablation_stages}
  \setlength{\tabcolsep}{3pt}
  \small 
  \begin{tabular}{l@{\hspace{3pt}}cc@{\hspace{5pt}}cc@{\hspace{5pt}}ccc}
      \toprule
      \multirow{2}{*}{Training Strategy} & \multicolumn{2}{c@{\hspace{5pt}}}{MAFW} & \multicolumn{2}{c@{\hspace{5pt}}}{MSP-IMPROV} & \multicolumn{3}{c}{Werewolf-XL} \\
      \cmidrule(lr){2-3} \cmidrule(lr){4-5} \cmidrule(lr){6-8}
      \space & UAR & WAR & UAR & WAR & A & V & D \\
      \midrule
      Stage~1 only & 43.07 & 57.69 & 65.53 & 75.67 & 35.51 & 69.48 & 43.91 \\
      \rowcolor{teal!20}
      Two-stage & \textbf{45.67} & \textbf{58.91} & \textbf{66.12} & \textbf{76.79} & \textbf{39.80} & \textbf{69.72} & \textbf{52.87} \\
      \rowcolor{teal!10}
      \multicolumn{1}{r}{$\Delta$\textit{Improvement}} & \textit{+2.60} & \textit{+1.22} & \textit{+0.59} & \textit{+1.12} & \textit{+4.29} & \textit{+0.24} & \textit{+8.96} \\
      Joint training & 41.25 & 56.43 & 64.87 & 74.92 & 33.82 & 68.95 & 41.73 \\
      \bottomrule
  \end{tabular}
\end{table}
First, we investigate the effectiveness of our two-stage training approach by comparing three training strategies: (1) Stage~1 only (contrastive learning and masked self-supervised learning), (2) our full two-stage approach (Stage~1 followed by Stage~2), and (3) joint training (applying both objectives simultaneously). As shown in Tab.~\ref{tab:ablation_stages}, the sequential two-stage approach consistently outperforms both alternatives. Joint training underperforms Stage~1-only training, suggesting that introducing MLLM knowledge without establishing strong cross-modal representations may confuse the learning process. This confirms our hypothesis that establishing general audio-visual correspondences before injecting emotion-specific knowledge creates a more effective foundation for emotion recognition. A significant finding is that even our Stage~1 model (without knowledge injection) already outperforms most existing approaches across datasets, as shown in Tab.~\ref{tab:ablation_stages}. The additional knowledge injection in Stage~2 provides consistent improvements, with gains of +2.60\% UAR on MAFW (from 43.07\% to 45.67\%) and +0.59\% UAR on MSP-IMPROV (from 65.53\% to 66.12\%). This confirms that our architecture with targeted MLLM knowledge integration effectively captures nuanced emotional cues that more complex models might miss. Furthermore, the t-SNE visualization in Fig.~\ref{fig:mafw_tsne} clearly shows how the knowledge injection in Stage~2 produces more distinct emotion clusters than the representations after Stage~1.

\noindent
\textbf{Analysis of Knowledge Injection Strategies.}
\begin{table}[t]
  \centering
  \caption{Ablation study on knowledge injection components on the MAFW dataset. LLM embeddings and Chain-of-Thought prompting yield the best performance.}
  \label{tab:ablation_knowledge}
  \begin{tabular}{lcc}
      \toprule
      \textbf{Knowledge Injection Strategy} & UAR & WAR \\
      \midrule
      \multicolumn{3}{l}{\textit{Embedding Type:}} \\
      \midrule
      CLIP embedding & 43.21 & 58.20 \\
      BERT embedding & 44.35 & 58.14 \\
      \rowcolor{teal!20}
      LLM embedding & \textbf{45.67} & \textbf{58.91} \\
      \midrule
      \multicolumn{3}{l}{\textit{Caption Generation:}} \\
      \midrule
      Basic MLLM prompt & 43.41 & 57.63 \\
      Emotion-focused prompt & 43.32 & 58.45 \\
      \rowcolor{teal!20}
      Chain-of-Thought prompt & \textbf{45.67} & \textbf{58.91} \\
      \bottomrule
  \end{tabular}
\end{table}
As described in Section 3.2, we use CoT prompting for caption generation and LLM for caption embedding. Next, we examine various aspects of our knowledge injection approach in Stage~2. Tab.~\ref{tab:ablation_knowledge} shows the impact of different embedding types, caption generation approaches, and contrastive learning objectives.
For embedding types, LLM-based embeddings outperform both CLIP embeddings and BERT embeddings, primarily because our captions exceed CLIP's 77-token length limit and contain specialized emotional vocabulary requiring the richer semantic understanding of LLMs. For caption generation, Chain-of-Thought prompting represents one of our key innovations, producing significantly richer emotion-relevant semantics than basic prompting approaches, while emotion-focused prompts offer moderate improvements over basic prompts, they lack the comprehensive emotional reasoning that CoT provides by first analyzing observable features before inferring emotional states.

\noindent
\textbf{Model Architecture Choices.} To better clarify our architectural design decisions, we compare single-stream versus dual-stream approaches. For fair comparison, all experiments were conducted using 100 epochs of Stage~1 training.
As shown in Tab.~\ref{tab:ablation_stream}, our single-stream architecture achieves comparable performance while using 45\% fewer parameters (39M vs. 71M). The dual-stream model essentially represents a simplified version of HiCMAE without hierarchical connections and cross-attention interactions. The single-stream model's efficiency comes from sharing parameters across modalities, creating a unified parameter space for audio and visual processing. The shared space enables tighter integration where features are processed through identical transformations. This reveals that for emotion recognition, parameter sharing from early processing stages can lead to better generalization, even with simpler fusion methods. 
Although computation costs are similar, the smaller parameter size makes single-stream model easier to deploy.
\begin{table}[t]
  \centering
  \caption{Comparison of dual-stream vs. single-stream architectures on MAFW dataset (Stage~1 training only).}
  \label{tab:ablation_stream}
  \resizebox{\columnwidth}{!}{%
  \begin{tabular}{lcccc}
      \toprule
      Architecture & UAR & WAR & \#Params (M) & FLOPs (G) \\
      \midrule
      Dual-stream & 43.01 & 56.67 & 71 & 49.33 \\
      \rowcolor{teal!20}
      Single-stream & \textbf{43.07} & \textbf{57.69} & \textbf{39} & 49.33 \\
      \bottomrule
  \end{tabular}%
  }
\end{table}

\begin{table}[t]
  \centering
  \caption{The impact of the weighting parameters ($\alpha$ and $\beta$) in Eq.~\eqref{eq:equation_5} for knowledge injection on the MAFW dataset.}
  \label{tab:ablation_hyperparams}
  \begin{tabular}{lcc}
      \toprule
      A-T vs. V-T weight: & UAR & WAR \\
      \midrule
      $\alpha=0.2, \beta=0.8$ & 44.32 & 57.85 \\
      $\alpha=0.4, \beta=0.6$ & 45.35 & 58.64 \\
      $\alpha=0.5, \beta=0.5$ & 45.20 & 58.42 \\
      \rowcolor{teal!20}
      $\alpha=0.6, \beta=0.4$ & \textbf{45.67} & \textbf{58.91} \\
      $\alpha=0.8, \beta=0.2$ & 44.95 & 58.42 \\
      \bottomrule
  \end{tabular}
\end{table}

\noindent
\textbf{Hyper-parameter Analysis.} Tab.~\ref{tab:ablation_hyperparams} shows the effect of different audio-text vs. video-text weights in Stage~2. Slightly favoring audio ($\alpha=0.6, \beta=0.4$) yields optimal performance, suggesting audio provides stronger emotional cues on MAFW. Extreme weighting in either direction reduces performance, confirming the importance of balanced modality integration during knowledge injection.

\section{Conclusion}

In this paper, we presented VAEmo, an efficient framework for audio-visual emotion representation learning that addresses the critical mismatch between general and emotion-specific representations. 
Through a carefully orchestrated two-stage pre-training approach on unlabeled audio-visual data, combined with our carefully designed emotional knowledge injection paradigm, our model effectively bridges the semantic gap between general audio-visual perception and emotion-specific understanding. Extensive experiments across multiple downstream benchmarks demonstrate that the 
VAEmo achieves state-of-the-art performance with strong generalization and a compact architecture.

\begin{acks}
This work was supported by the National Natural Science Foundation of China (NSFC, No. 62202139, No. 62172138) and the Fundamental Research Funds for the Central Universities (No. JZ2025HGTB0226, No. JZ2024HGTG0310). The computation is completed on the HPC Platform of Hefei University of Technology.
\end{acks}

\bibliographystyle{ACM-Reference-Format}
\balance
\bibliography{VAEmo_reference}

\end{document}